\documentclass[sigconf]{acmart}

\usepackage{subfig}
\usepackage{graphicx}
\usepackage{float}
\usepackage{array}
\usepackage{algorithm}
\usepackage{algorithmic}
\usepackage{multirow}
\usepackage{colortbl}
\usepackage{hyperref}
\definecolor{gray}{RGB}{230,230,230}
\definecolor{lightyellow}{RGB}{255, 249, 192}
\definecolor{lightpurple}{RGB}{248, 231, 242}
\definecolor{timecolor}{RGB}{0, 115, 140}
\newcommand{\lat}[1]{\textcolor{timecolor}{#1}}

\usepackage{amsmath,amsfonts,bm}

\def\eqref#1{equation~\ref{#1}}

\def\1{\bm{1}}

\def\vp{{\bm{p}}}

\def\vy{{\bm{y}}}
\def\vz{{\bm{z}}}

\DeclareMathAlphabet{\mathsfit}{\encodingdefault}{\sfdefault}{m}{sl}
\SetMathAlphabet{\mathsfit}{bold}{\encodingdefault}{\sfdefault}{bx}{n}

\newcommand{\softmax}{\mathrm{softmax}}

\usepackage{hyperref}
\hypersetup{
    colorlinks=true,
    urlcolor=magenta, 
    linkcolor=blue,
    citecolor=green
}
\AtBeginDocument{%
  \providecommand\BibTeX{{%
    \normalfont B\kern-0.5em{\scshape i\kern-0.25em b}\kern-0.8em\TeX}}}
\renewcommand\footnotetextcopyrightpermission[1]{}
\setcopyright{none}
\copyrightyear{2026}
\acmYear{2026}

\acmConference[MM '26]
{The 34th ACM International Conference on Multimedia}
{November 10--14, 2026}
{Rio de Janeiro, Brazil}

\acmBooktitle{Proceedings of the 34th ACM International Conference on Multimedia
(MM '26), November 10--14, 2026, Rio de Janeiro, Brazil}

\begin{document}

\title{Continual Video-MLLM Adaptation over Evolving Domains}
\author{Rui Cheng$^{1,2}$, Meixing Shi$^{1,2}$, Yuxiang Cai$^{1,2*}$, Jingcai Guo$^{3}$, Jianwei Yin$^{1,2}$, Zhi Chen$^{4*}$ \\
$^{1}$ School of Software Technology, Zhejiang University, Ningbo, China \\
$^{2}$ Zhejiang Key Laboratory of Digital-Intelligence Service Technology, China \\
$^{3}$ Hong Kong Polytechnic University, Hong Kong SAR \\
$^{4}$ The University of Southern Queensland, Australia \\
uqzhichen@gmail.com, caiyuxiang@zju.edu.cn \\
\texttt{\url{https://github.com/ChengRui310/DAER}}
}\authornote{Zhi Chen and Yuxiang Cai are the corresponding authors.}













\renewcommand{\shortauthors}{Rui Cheng et al.}

\begin{abstract}
Video multimodal large language models (Video-MLLMs) have shown strong capability in video understanding, yet their adaptation to sequentially evolving domains remains underexplored. In real-world deployments, video data often arrives continuously from heterogeneous domains, requiring the model to acquire new domain-specific knowledge without overwriting previously learned capabilities. Existing continual learning methods typically rely on shared adaptation spaces, which can induce severe cross-domain interference and catastrophic forgetting. We propose Distribution-Aware Expert Routing (DAER), a parameter-efficient framework for continual Video-MLLM adaptation over evolving domains. DAER maintains domain-isolated lightweight experts while keeping the pretrained Video-MLLM backbone frozen, thereby decoupling domain-specific adaptation from the general multimodal knowledge of the pretrained model. To enable fine-grained specialization, we introduce an intra-domain distribution-aware routing mechanism that matches each input to expert-level prototype reservoirs using Maximum Mean Discrepancy (MMD). To address the absence of task identities at inference time, we further propose an inter-domain routing mechanism that performs prototype matching in a discriminative subspace for robust domain identification. In addition, we introduce adaptive domain merging to improve parameter scalability and adopt a two-stage optimization strategy to stabilize expert specialization during continual learning. We evaluate DAER by curating a domain-incremental benchmark built from ten VidQA datasets covering diverse visual environments and reasoning demands. Experiments on two strong Video-MLLM backbones show that DAER consistently outperforms prior methods. On InternVideo2.5-8B, for example, DAER improves the average accuracy from $64.38\%$ to $67.59\%$ over the strongest prior baseline while reducing backward forgetting from $1.81$ to $-0.14$. These results demonstrate that combining domain-isolated expert adaptation with distribution-aware routing provides an effective and scalable solution for continual Video-MLLM adaptation under evolving domains. 

\end{abstract}

\maketitle

\section{Introduction}

Video multimodal large language models (Video-MLLMs)~\cite{zeng2024timesuite,wang2025internvideo2,zhang2025videollama,zhang2024llava} have recently demonstrated strong capability in understanding dynamic visual content and natural language queries. By combining powerful visual encoders with large language models, these systems have substantially advanced video understanding tasks such as video question answering.
Despite this progress, most existing Video-MLLMs are developed under the conventional offline learning paradigm, where all training data are assumed to be available simultaneously. This assumption is often unrealistic in real-world deployments, where data arrive sequentially from evolving domains, such as daily-life videos, traffic scenes, educational cartoons, and abstract reasoning environments.

\begin{figure}[t]
    \centering
    \includegraphics[width=0.99\linewidth]{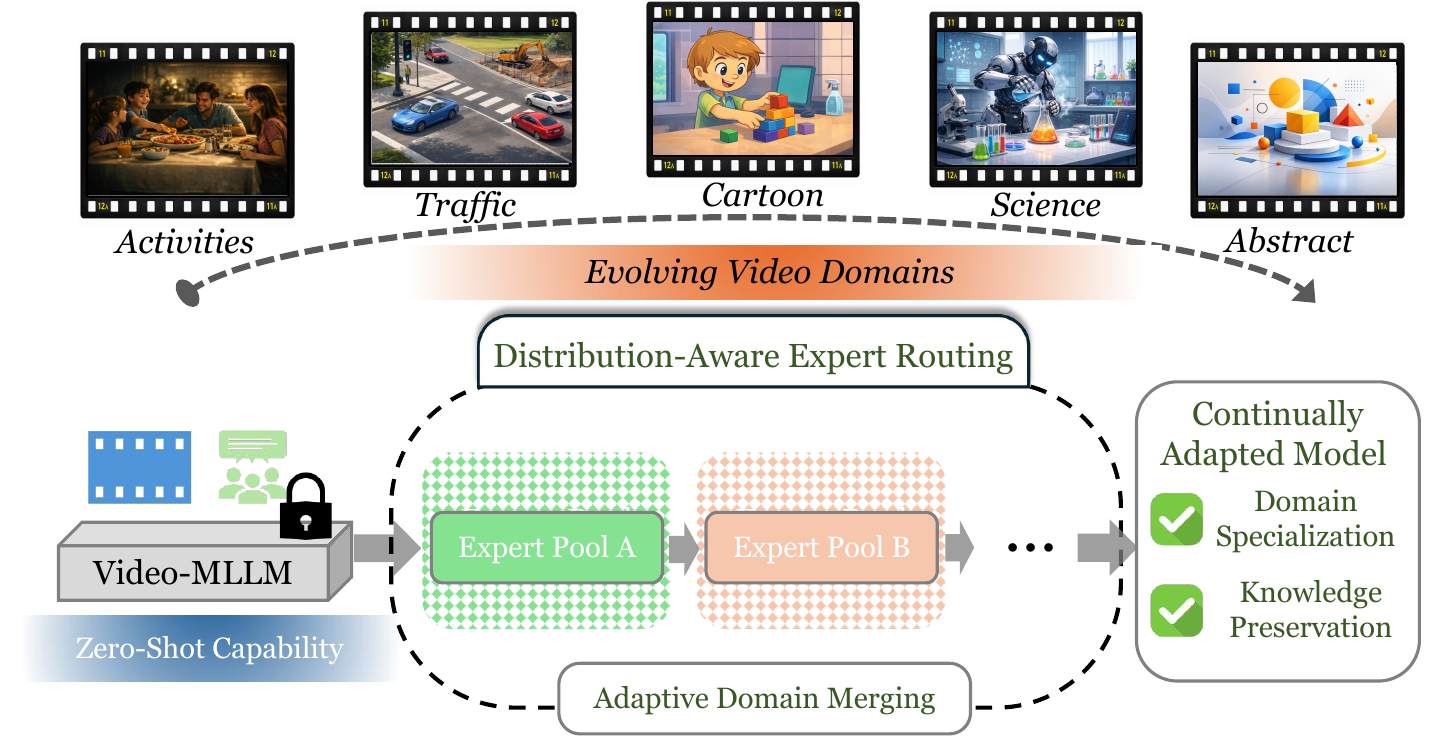}
\caption{Overview of DAER for continual Video-MLLM adaptation over evolving domains. DAER routes inputs from sequentially arriving video domains to domain-specific expert pools. Moreover  similar domains can be adaptively merged for improved parameter efficiency.}
    \label{fig:thumbnail}
\end{figure}

Continual learning~\cite{kirkpatrick2017overcoming,rebuffi2017icarl,riemer2018learning,hadsell2020embracing} offers a natural paradigm for enabling long-term adaptation in such non-stationary settings. In particular, domain-incremental continual learning~\cite{lamers2023clustering,dai2023domain,mirza2022efficient} considers the practical scenario in which the learning objective and output space remain unchanged, while the underlying data distribution shifts over time. This setting is especially relevant to Video-MLLMs, whose downstream applications often span heterogeneous domains with substantial differences in visual style, scene composition, motion patterns, question semantics, and reasoning demands. A continually adapted model must therefore acquire new domain-specific knowledge while preserving previously learned capabilities. However, directly fine-tuning a single model across sequential domains often leads to severe catastrophic forgetting, whereas naively sharing adaptation parameters across domains can introduce strong cross-domain interference.

Existing continual learning strategies only partially address this challenge. Regularization-based methods \cite{li2017learning,kirkpatrick2017overcoming,zenke2017continual} constrain parameter updates to preserve prior knowledge, but often struggle under large domain shifts. Replay-based \cite{lopez2017gradient,cha2021co2l} methods alleviate forgetting by revisiting previous samples, yet they introduce additional memory cost and may be undesirable in privacy-sensitive settings. Prompt-based \cite{wang2022learning,wang2022dualprompt,smith2023coda,cai2024empowering} and adapter-based methods \cite{yu2024boosting,cheng2024dam} improve parameter efficiency, but they still commonly rely on shared adaptation spaces or soft parameter partitioning, which may be insufficient to prevent interference among highly heterogeneous domains. For continual Video-MLLM adaptation over evolving domains, recent approaches such as dynamic adapter merging \cite{cheng2024dam} and collaborative prompting \cite{cai2024empowering} have shown encouraging progress, but robust domain-specific adaptation without explicit task identity while largely preserving the general capabilities of the pretrained Video-MLLM remains an open problem.

In this work, as shown in Fig. \ref{fig:thumbnail}, we propose Distribution-Aware Expert Routing (\textbf{DAER}), a parameter-efficient framework for continual Video-MLLM adaptation over evolving domains. Our key idea is to combine strict parameter isolation with distribution-aware routing. Specifically, instead of forcing all domains to update a shared adaptation space, we maintain task-specific pools of lightweight LoRA-style experts while keeping the pretrained Video-MLLM backbone frozen. This design preserves the general multimodal knowledge, while confining incremental learning to domain-specialized adaptation modules. To further enable fine-grained specialization within each task, we introduce an intra-domain routing mechanism, which measures the similarity between the current input representation and expert-level prototype reservoirs. Compared with conventional parametric gating \cite{shazeer2017outrageously}, this non-parametric routing strategy provides a more direct way to activate experts based on the underlying domain distribution.

A practical difficulty in continual inference is that explicit task identifiers are unavailable at test time. To address this issue, we further propose an inter-domain distribution-aware routing mechanism for task prediction. Instead of matching in the raw feature space, we construct a discriminative subspace for prototype matching, enabling more robust domain identification under overlapping feature distributions. Based on the inferred task, the model activates the corresponding expert pool. In addition, since strict parameter isolation causes the number of expert pools to grow with the task stream, we further introduce an adaptive domain merging strategy that selectively merges similar historical domains according to prototype-level distribution discrepancy.

To stabilize learning, we adopt a two-stage optimization strategy. We first perform a warm-up stage to initialize prototype reservoirs and establish a stable routing space, and then activate the full DAER routing mechanism for end-to-end optimization. This staged design alleviates the cold-start instability of expert prototypes and enables more reliable expert specialization during continual adaptation.

We evaluate the proposed framework on a domain-incremental benchmark built from ten VidQA datasets spanning realistic scenes, traffic understanding, physical reasoning, story understanding, humor, creativity, and abstraction. Experiments on two strong Video-MLLM backbones, including InternVideo2.5 and VideoLLaMA3, show that DAER consistently outperforms prior methods. On InternVideo2.5-8B, for example, DAER improves the average accuracy from $64.38\%$ to $67.59\%$ over the strongest prior baseline while reducing backward forgetting from $1.81$ to $-0.14$. These results demonstrate that combining domain-isolated expert adaptation with distribution-aware routing yields a strong and stable solution for continual Video-MLLM adaptation over evolving domains.

Our contributions are summarized as follows:
\begin{itemize}
    \item We propose DAER, a parameter-efficient continual adaptation framework that combines a frozen Video-MLLM backbone with domain-isolated lightweight expert pools, thereby reducing cross-domain interference while preserving the zero-shot capability of the pretrained model.
    \item We introduce a distribution-aware routing design that includes MMD-based intra-domain expert routing and discriminative inter-domain prototype matching for task prediction. We further develop adaptive domain merging and a two-stage optimization strategy to improve parameter scalability and stabilize expert specialization during continual learning.
    \item Extensive experiments on ten VidQA datasets and two strong backbones demonstrate that the proposed method achieves state-of-the-art performance in domain-incremental continual VidQA.
\end{itemize}

\section{Related Work}

\noindent \textbf{Continual Learning for Video-Language Understanding.} 
Continual learning (CL)~\cite{kirkpatrick2017overcoming,rebuffi2017icarl,riemer2018learning,hadsell2020embracing,zhao2025continual,wang2025discrimination} aims to enable models to learn from a sequential stream of tasks while mitigating catastrophic forgetting~\cite{mcclelland1995there} of previously acquired knowledge. Depending on how the data distribution evolves over time, CL is commonly studied under class-incremental \cite{van2022three,masana2022class,zhou2024class} and domain-incremental \cite{van2022three,shi2023unified} settings. Early work mainly focused on image classification, with methods broadly categorized as regularization-based~\cite{li2017learning,kirkpatrick2017overcoming,zenke2017continual}, replay-based~\cite{cha2021co2l,lopez2017gradient}, and prompt-based 
approaches~\cite{wang2022learning,wang2022dualprompt,wang2022s,smith2023coda,cai2024empowering}. Compared with image CL, continual learning for video and video-language understanding remains relatively underexplored. Recent works have begun to address this setting from different perspectives. For domain-incremental continual VidQA, Cheng et al.~\cite{cheng2024dam} learn domain-specific adapters and dynamically merge them at inference time, while Cai et al.~\cite{cai2024empowering} use collaborative prompting to model question semantics, video content, and temporal dynamics. Beyond VidQA, Villa et al.~\cite{villa2023pivot} study class-incremental action recognition with pretrained vision-language models, and Tang et al.~\cite{tang2024vilco} introduce ViLCo-Bench and a query-incremental setting for broader video-language continual learning. Mixture-of-Experts (MoE)~\cite{jacobs1991adaptive} has also emerged as a promising direction, since activating only a subset of experts can reduce cross-task interference while preserving capacity. Representative works include expert-based continual learning~\cite{aljundi2017expert}, theoretical analyses of MoE in continual learning~\cite{li2024theory}, and incremental MoE adapters for vision-language models~\cite{yu2024boosting}. Motivated by these advances, our work explores a distribution-aware MoE framework for continual Video-MLLM adaptation, where lightweight experts are routed according to distribution similarity to enable stable specialization under evolving domains.

\noindent \textbf{Video-MLLMs and Video Question Answering.} \hspace{3pt}
Recent advances in large language models (LLMs)~\cite{brown2020language,chowdhery2023palm,touvron2023llama} and vision-language models~\cite{liu2023visual,alayrac2022flamingo,li2023blip,wei2026benchmarking,chen2026spectral,chen2026accelerating,zhao2026generative,chen2025cluster,zhao2025synthetic,wang2026learning} have enabled Video-MLLMs or video-language models (VLMs)~\cite{zeng2024timesuite,wang2025internvideo2,zhang2025videollama,zhang2024llava}, which extend multimodal understanding from images to videos through joint visual-text modeling. Compared with static images, videos contain richer temporal structure and dynamic semantics, making temporal modeling and cross-modal reasoning more challenging. Representative Video-MLLMs, such as LLaVA-Video~\cite{zhang2024llava}, VideoLLaMA3~\cite{zhang2025videollama}, and InternVideo2.5~\cite{wang2025internvideo2}, adopt different strategies to improve temporal understanding and computational efficiency, demonstrating strong video understanding capabilities across diverse scenarios. VidQA is a core capability of Video-MLLMs, requiring joint reasoning over dynamic visual content and natural language questions. Transformer-based architectures dominate this task~\cite{yang2022zero,wang2025internvideo2,lei2021less,cheng2023vindlu,xiao2022video,zhang2025videollama}, with models such as FrozenBiLM~\cite{yang2022zero} and InternVideo2.5~\cite{wang2025internvideo2} showing strong performance. Despite this progress, most Video-MLLMs and VidQA methods are still developed under the offline learning assumption. In contrast, our work focuses on continual adaptation over evolving domains, with VidQA serving as a representative and challenging testbed for studying continual Video-MLLM adaptation.

\begin{figure*}[!t]
    \centering
    \includegraphics[width=1.0\linewidth]{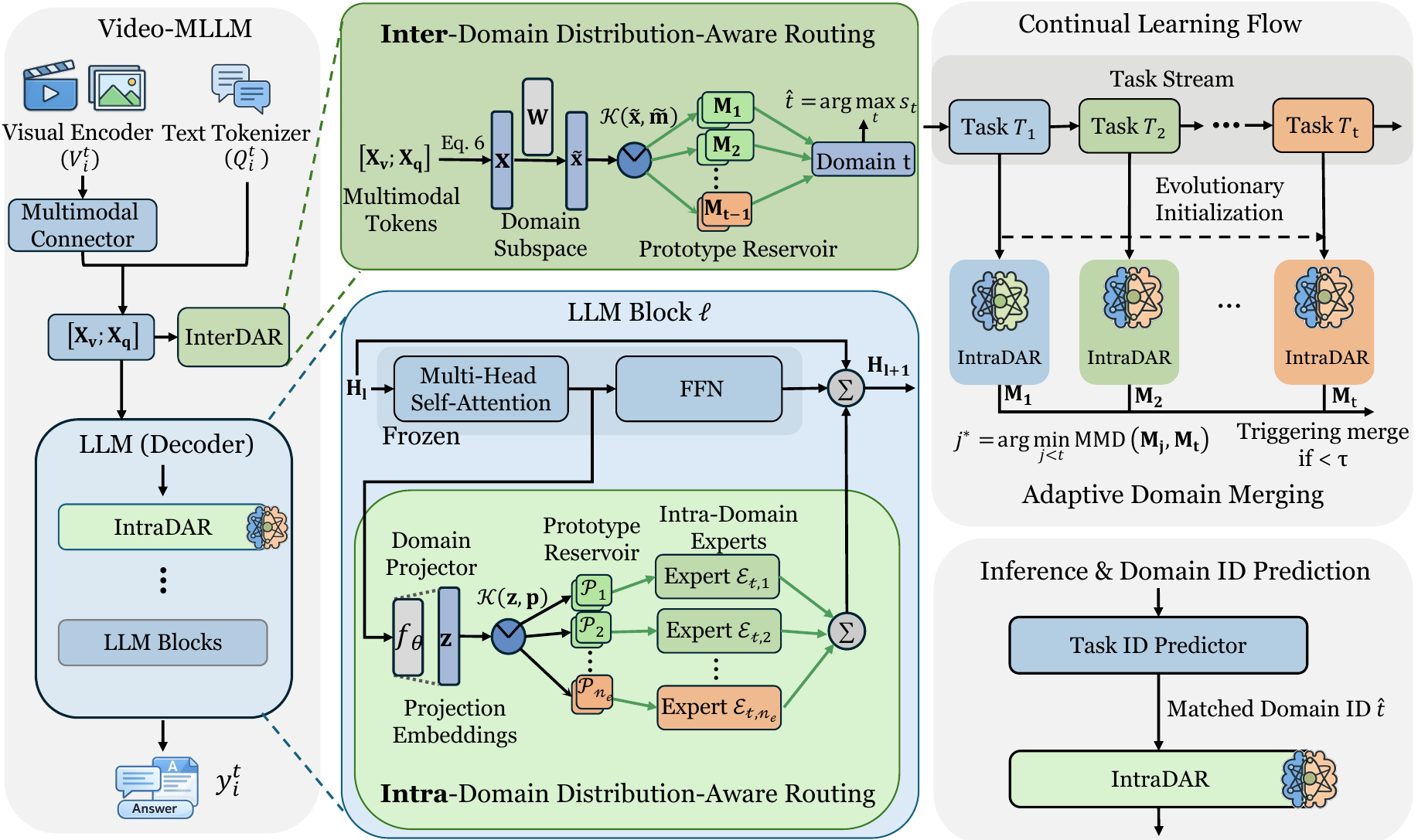}
    \caption{Illustration of the DAER framework for continual Video-MLLM adaptation over evolving domains. DAER performs inter-domain routing to identify the most relevant domain, intra-domain routing to activate domain-specific experts and adaptive domain merging to reduce redundancy across similar domains.}
    \label{fig:arc}
\end{figure*}

\section{Methodology}

\subsection{Preliminaries}
\noindent \textbf{Continual Video-MLLM Adaptation over Evolving Domains.}
In this continual learning setting, a model is trained sequentially on a series of tasks $t = 1, \ldots, T$. Each task $t$ is associated with a dataset $\mathcal{D}_t = \{(V_i^t, Q_i^t, y_i^t)\}_{i=1}^{n_t}$, where $V_i^t$ denotes the video input, ${Q}_i^t$ is the question text, and $y_i^t$ represents the target answer sequence. 
Each dataset is drawn from a task-specific joint distribution $\mathcal{D}_t \sim P_t(V, Q, Y)$. 
We can only access $\mathcal{D}_t$ when training on task $t$.
The architecture of a Video-MLLM typically contains a visual encoder, a multimodal connector, and a LLM decoder. Specifically, we first sample key frames from the input video $V_i^t$ and a visual backbone is utilized to extract $n_v$ visual tokens $\mathbf{X}_v \in \mathbb{R}^{n_v \times d}$. Concurrently, the question text $Q_i^t$ is
tokenized as $n_q$ text embeddings $\mathbf{X}_q \in \mathbb{R}^{n_q \times d}$. A multimodal connector aligns the visual features with the LLM's embedding space. We then concatenate these tokens into a unified input sequence:
$$\mathbf{H}_0 = [\mathbf{X}_v; \mathbf{X}_q] \in \mathbb{R}^{(n_v+n_q) \times d},$$
where $\mathbf{H}_0$ serves as the initial hidden state for the LLM. The model predicts each token $y_j$ in the answer sequence $y$ in an autoregressive manner:
$$P(y_j | \mathbf{H}_0, y_{<j}) = \text{Softmax}(\text{LLM}(\mathbf{H}_0, y_{<j})).$$
Given the continuously shifting domain distributions, the primary challenge is to achieve domain-specific adaptation without compromising the structural stability of multimodal alignments. 

\noindent \textbf{Maximum Mean Discrepancy (MMD).}
MMD is a non-parametric metric for measuring the discrepancy between two probability distributions in a reproducing kernel Hilbert space (RKHS). Given two distributions $\rho$ and $\sigma$, MMD compares their mean embeddings in the kernel-induced feature space. Formally, the squared MMD is defined as:
\begin{equation}
\mathrm{MMD}^2(\rho,\sigma) = \left\| \mu_\rho - \mu_\sigma \right\|_{\mathcal{H}}^2,
\end{equation}
where $\mu_\rho = \mathbb{E}_{x \sim \rho}[\phi(x)]$ and $\mu_\sigma = \mathbb{E}_{y \sim \sigma}[\phi(y)]$ denote the mean embeddings of $\rho$ and $\sigma$ in the RKHS $\mathcal{H}$, and $\phi(\cdot)$ is the kernel feature mapping. 
In practice, since the underlying distributions are unknown, MMD is estimated from finite samples using pairwise kernel evaluations. 
Given two finite sample sets $X = \{x_i\}_{i=1}^n$ and $Y = \{y_j\}_{j=1}^m$ drawn from distributions $\rho$ and $\sigma$, respectively, we adopt the empirical estimator to compute the squared MMD distance:
\begin{equation}
\begin{split}
\mathrm{MMD}^{2}(X, Y) 
&= \frac{1}{n^{2}} \sum_{i,j=1}^{n} \mathcal{K}(x_{i}, x_{j})
+ \frac{1}{m^{2}} \sum_{i,j=1}^{m} \mathcal{K}(y_{i}, y_{j}) \\
&- \frac{2}{nm} \sum_{i=1}^{n} \sum_{j=1}^{m} \mathcal{K}(x_{i}, y_{j})    
\end{split}
\label{MMDeq}
\end{equation}
where $\mathcal{K}(\cdot, \cdot)$ denotes the kernel function.
A smaller MMD value indicates that the two samples are drawn from more similar distributions. In our setting, this property makes MMD particularly suitable for measuring the similarity between the current input distribution and the distribution represented by each expert, thereby enabling distribution-aware expert routing.

\subsection{Distribution-Aware Expert Routing (DAER)}
To overcome the cross-task interference caused by parameter sharing in continual Video-MLLM adaptation, we propose DAER. The core idea is to ensure strict parameter independence through domain-isolated expert pools, while employing a non-parametric MMD-based routing mechanism to precisely activate domain-specific states according to the underlying data distribution. Throughout the continual learning process, we keep the backbone parameters of the Video-MLLM frozen, thereby fully preserving the general knowledge of the pretrained model, while confining all domain-specific knowledge to DAER.
As illustrated in Figure~\ref{fig:arc}, we integrate the DAER into each layer of the LLM. For the $l$-th Transformer layer, let $\mathbf{H}_l$ denote its input token sequence. 
After multi-head self-attention (MHSA), we obtain a hidden representation $\hat{\mathbf{H}}_l = \text{MHSA}(\mathbf{H}_l)$.  
On top of this, we introduce a LoRA-based \cite{hu2022lora} domain-isolated MoE adaptation module in parallel to the FFN branch. 
This module takes $\hat{\mathbf{H}}_l$ as input, performs low-rank residual updates, and dynamically activates the corresponding experts. 
For the task $t$, the final output representation of the $l$-th layer, $\mathbf{H}_{l+1}$ is computed as: 
\begin{equation}
\mathbf{H}_{l+1} 
= \hat{\mathbf{H}}_l 
+ \mathrm{FFN}(\hat{\mathbf{H}}_l) 
+ \mathrm{MoE}_t(\hat{\mathbf{H}}_l).
\end{equation}
Specifically, we allocate a dedicated set of lightweight experts $\{\mathcal{E}_{t,1}, \dots, \mathcal{E}_{t,n_e}\}$ for each task $t$, where $n_e$ is the number of experts. During the learning of the current task $t$, all task-specific parameters from previous tasks $\{\mathrm{MoE}_1, \dots, \mathrm{MoE}_{t-1}\}$ remain frozen unless a similar previous domain is explored (\S {Adaptive Domain Merging}). 
This physical parameter isolation mechanism ensures that the gradient updates for task $t$ do not cause any shifts in the parameter space of older tasks, thereby fundamentally addressing the issue of catastrophic forgetting from a structural perspective.

\vspace{3pt}
\noindent \textbf{Intra-Domain Distribution-Aware Routing (IntraDAR).}
To enable more fine-grained expert invocation within each domain, we introduce an intra-domain routing mechanism based on MMD. 
Unlike conventional MoE architectures that rely on a parameterized linear layer for discriminative routing, we hypothesize that experts whose modeled distributions better match the current input will achieve superior modeling performance. 
Based on this assumption, we adopt MMD as a distributional similarity measure for expert selection. 
We employ a domain projector $f_{\theta}: \mathbb{R}^{(n_v+n_q)\times d} \rightarrow \mathbb{R}^{d}$ to map the hidden state $\hat{\mathbf{H}}_l$ of the LLM into a domain representation space, yielding $\vz = f_{\theta}(\hat{\mathbf{H}}_l ) \in \mathbb{R}^{d}$. For $i_e$-th expert $\mathcal{E}_{t,i_e}$ in task $t$, we maintain an expert-level prototype reservoir $\mathcal{P}_{i_e} = \{\vp_{i_p}\}_{i_p=1}^{n_p} \in \mathbb{R}^{n_p \times d}$, where $n_p$ is the number of the maintained prototypes. 
According to Eq.~(\ref{MMDeq}), under the current problem setting, the first term depends only on the current input representation $\vz$ and is thus constant across all experts; the second term characterizes the internal self-similarity of the expert-level prototype reservoir $\mathcal{P}$, and under the assumption that the prototype distribution remains relatively stable, its influence on routing decisions is relatively limited. Therefore, under this approximation, minimizing the MMD distance between the sample and the expert distribution can be approximately reduced to maximizing the cross-kernel similarity between them. 
Accordingly, we define the routing signal to the $i_e$-th expert as:
\begin{equation}
s_{i_e}(\vz) = \frac{1}{n_p} \sum_{\vp \in \mathcal{P}_{i_e}} \mathcal{K}(\vz, \vp),
\end{equation}
where $\mathcal{K}(\cdot, \cdot)$ denotes a multi-kernel Gaussian (RBF) 
function to enhance distributional discrimination capability. A larger 
routing score indicates that the input sample is closer to the distribution 
modeled by the expert in the domain space, and thus is more likely to benefit 
from that expert’s modeling capacity.
Finally, we select the Top-$k$ experts with the highest routing scores and 
aggregate their outputs to produce the final MoE output:
\begin{equation}
\vy 
= \sum_{i \in \mathcal{S}} 
\alpha_i(\vz) \, \mathcal{E}_i(\hat{\mathbf{H}}_l), \hspace{5pt}
\mathcal{S} = \operatorname{Top}k \left(
\softmax \left( \{ s_j(\vz) / \tau \}_{j=1}^{n_e} \right)
\right), 
\end{equation}
where $k < n_e$, $\mathcal{S}$ denotes the index set of activated experts, and the routing weights $\alpha_i(\vz)$ are computed with Softmax on $s(\vz)$.
For each input sample with domain representation $\vz$, we first identify the expert with the highest routing score, i.e.,
$
i^* = \arg\max_{i_e \in \{1,\dots,n_e\}} s_{i_e}(\vz),
$
and then assign $\vz$ to the corresponding prototype reservoir $\mathcal{P}_{i^*}$ so as to continuously refine the local distribution characterized by that expert. When $|\mathcal{P}_{i^*}| < n_p$, the new domain representation $\vz$ is directly inserted into $\mathcal{P}_{i^*}$; otherwise, the reservoir is dynamically updated via reservoir sampling, such that the maintained prototype set remains representative of the local domain distribution modeled by expert $\mathcal{E}_{t,i^*}$.

\vspace{3pt}
\noindent \textbf{Inter-Domain Distribution-Aware Routing (InterDAR).} \hspace{5pt}
In continual learning, tasks of different domains arrive sequentially in a streaming manner without explicit domain or task identifiers. Therefore, during inference, the model must automatically determine the target domain of the current input~\cite{abati2020conditional,henning2021posterior,lin2023class}. To this end, we propose an inter-domain distribution-aware routing mechanism. Specifically, for each task $t$, we maintain a domain-level prototype reservoir $\mathbf{M}_t = \{\mathbf{m}_{t,i_m}\}_{i_m=1}^{n_m}$ during training, where $n_m$ is the number of prototypes per task. This prototype reservoir serves as a finite-sample approximation of the underlying task distribution. Each prototype $\mathbf{m}_{t,i_m}$ is obtained by aggregating the textual semantic representation $\mathbf{X}_q$ and the video representations $\mathbf{X}_v$ with ratio $\alpha$:
\begin{equation}
\mathbf{m}_{t,i_m} = \alpha \cdot \frac{1}{n_q}\sum_{i_q=1}^{n_q}\mathbf{X}_q + (1-\alpha) \cdot
\frac{1}{n_v} \sum_{i_v=1}^{n_v} \mathbf{X}_v.
\label{eqPro}
\end{equation}

To improve task recognition under overlapping domain distributions, we perform prototype matching in a distribution-aware discriminative subspace rather than in the original feature space. Specifically, we adopt a two-stage projection strategy that first regularizes the covariance structure of the feature space by removing redundant low-variance directions $\mathbf{W}_{\mathrm{reg}} \in \mathbb{R}^{d \times r}$, and then learns a supervised discriminative subspace that enlarges inter-domain margins while suppressing intra-task variation $\mathbf{W}_{\mathrm{disc}} \in \mathbb{R}^{r \times (t-1)}$.
We denote the resulting transformation by $\mathbf{W} = \mathbf{W}_{\mathrm{reg}} \mathbf{W}_{\mathrm{disc}} \in \mathbb{R}^{d \times (t-1)}$, where $\mathbf{W}_{\mathrm{reg}}$ performs feature-space regularization and $\mathbf{W}_{\mathrm{disc}}$ captures domain-discriminative directions.
Thus, given an input feature $\mathbf{x}$ obtained also using Eq. \ref{eqPro} and the $i_m$-th prototype $\mathbf{m}_{t,i_m}$ of task $t$, we project them into the discriminative subspace:
\begin{equation}
\tilde{\mathbf{x}} = \mathbf{x}\mathbf{W}, \qquad
\tilde{\mathbf{m}}_{t,i_m} = \mathbf{m}_{t,i_m}\mathbf{W}.
\end{equation}
We then evaluate their affinity using a kernel function defined in the projected space.
The task matching score is obtained by aggregating the prototype affinities of task $t$:
\begin{equation}
s_t = \frac{1}{n_m}\sum_{i_m=1}^{n_m}\mathcal{K}(\tilde{\mathbf{x}}, \tilde{\mathbf{m}}_{t,i_m}),
\end{equation}
where $n_m$ denotes the number of prototypes associated with task $t$. The predicted task is then given by $\hat{t} = \arg\max_t s_t$.

\begin{table*}[t]
\centering
\caption{Comparison of different methods under the continual learning setting based on InternVideo2.5-8B.}
\setlength{\tabcolsep}{4.8pt}
\begin{tabular}{lcccccccccccc}
\toprule
\multicolumn{13}{l}{\textbf{Training Order:} MSVD $\rightarrow$ MSRVTT $\rightarrow$ Traffic $\rightarrow$ Pororo $\rightarrow$ SVQA $\rightarrow$ Physics $\rightarrow$ Magic $\rightarrow$ Humor $\rightarrow$ Creative $\rightarrow$ Abstraction} \\
\midrule
\multirow{2}{*}{Method} & \multicolumn{10}{c}{Downstream VidQA Accuracy (\%)} & \multirow{2}{*}{Avg.} & \multirow{2}{*}{\lat{BWF$\downarrow$}}  \\
\cmidrule(lr){2-11}
& MSVD & MSRVTT & Traffic & Pororo & SVQA & Physics & Magic & Humor & Creative & Abstraction & \\
\midrule
Zero-Shot 
& 55.50 & 40.30 & 54.08 & 54.10 & 53.67 & 65.98 & 63.70 & 64.04 & 65.08 & 55.19 & 57.16 & -- \\
Seq-FT 
& 55.90 & 44.40 & 63.04 & 57.58 & 65.83 & 67.22 & 67.51 & 67.69 & 66.18 & 60.90 & 61.63 & -- \\
\midrule

\rowcolor{lightyellow} \multicolumn{13}{l}{\textit{Upper Bound (Multi-task Finetuning)}} \\
Adapters 
& 60.80 & 51.00 & 67.32 & 65.58 & 69.67 & 70.82 & 71.43 & 71.54 & 69.65 & 62.28 & 66.00 & -- \\
Prompt Tuning 
& 61.90 & 48.00 & 67.40 & 62.03 & 69.67 & 68.19 & 69.84 & 66.92 & 69.47 & 59.00 & 64.24 & -- \\
\midrule

\rowcolor{lightyellow} \multicolumn{13}{l}{\textit{Regularization-based Methods}} \\
LwF 
& 55.55 & 42.45 & 59.52 & 57.23 & 65.75 & 68.47 & 67.72 & 66.92 & 67.28 & 56.40 & 60.73 & \lat{4.24} \\
EWC 
& 55.30 & 44.20 & 61.68 & 56.61 & 66.75 & 69.43 & 67.41 & 67.60 & 67.64 & 60.90 & 61.75 & \lat{4.18} \\
\midrule

\rowcolor{lightyellow} \multicolumn{13}{l}{\textit{Prompt-based Methods}} \\
L2P 
& 57.80 & 45.00 & 57.96 & 56.75 & 65.08 & 63.62 & 67.83 & 65.58 & 67.46 & 58.82 & 60.59 & \lat{2.48} \\
CODA-Prompt 
& 58.45 & 45.80 & 61.44 & 58.76 & 64.42 & 63.35 & 66.77 & 66.35 & 67.28 & 57.96 & 61.06 & \lat{3.13} \\
S-Prompts 
& 60.75 & 45.85 & 67.64 & 62.17 & 70.17 & 64.45 & 68.25 & 67.02 & 68.19 & 56.92 & 63.14 & \lat{\underline{0.26}} \\
\midrule

\rowcolor{lightyellow} \multicolumn{13}{l}{\textit{Adapter-based Methods}} \\
MoE-Adapters 
& 58.05 & 47.75 & 63.12 & 55.42 & 65.92 & 64.45 & 67.83 & 67.12 & 65.26 & 60.73 & 61.57 & \lat{3.43} \\
DAM 
& 59.95 & 48.05 & 67.84 & 64.95 & 71.25 & 67.36 & 68.89 & 68.56 & 70.02 & 56.92 & \underline{64.38} & \lat{1.81} \\
\midrule
\rowcolor{lightpurple} \textbf{DAER}~(Ours)
& \textbf{62.00} & \textbf{48.85} & \textbf{69.84} & \textbf{68.22} & \textbf{77.50} & \textbf{70.68} & \textbf{72.59}& \textbf{70.48} & \textbf{73.31} & \textbf{62.46} & \textbf{67.59} & \lat{\textbf{-0.14}} \\
\bottomrule
\end{tabular}
\label{tab:main_results}
\end{table*}

\vspace{3pt}
\noindent \textbf{Adaptive Domain Merging (ADM).}  \hspace{5pt}
Although we can achieve strong performance in continual learning by isolating task-specific parameters, it introduces an independent MoE for each task, making the parameter size increase continuously as the number of tasks grows. To mitigate parameter redundancy, we further propose an adaptive domain merging mechanism. Specifically, the distribution discrepancy between tasks is quantified by MMD between the domain-level prototype reservoirs. For the current task $t$, we first identify the most similar historical task according to:
\begin{equation}
j^*=\arg\min_{j<t}\mathrm{MMD}\!\left(\mathbf{M}_j,\mathbf{M}_t\right).
\end{equation}
Domain merging is triggered when the MMD-based domain discrepancy between two domains falls below the threshold $\tau$. In this case, the expert parameters of the current task are averaged with those of the most similar historical task; otherwise, the expert parameters of the current task are preserved.

\vspace{3pt}
\noindent \textbf{Optimization.} \hspace{5pt}
To alleviate the training instability caused by the cold-start issue of expert prototype reservoir at the early stage of training, we adopt a two-stage training strategy:
\textbf{1) Warm-up.} 
At the beginning, the domain prototype reservoir is not yet established. Directly enabling routing may lead to unreliable distribution estimation. To achieve a uniform initialization of the prototype reservoir, we introduce an auxiliary gating network $g_\phi$ implemented as a linear layer. During this stage, we only optimize the domain router $f_\theta$ and the auxiliary gating network $g_\phi$. The auxiliary gate produces smooth cross-expert soft assignment probabilities via a high-temperature Softmax:
\begin{equation}
w_k(\mathbf{z})
=
\frac{
\exp\big(g_\phi(\vz)_k / \tau_{\text{warm}}\big)
}{
\sum_{j=1}^{n_e}
\exp\big(g_\phi(\vz)_j / \tau_{\text{warm}}\big)
}.
\end{equation}
The domain embedding $\vz$ is fed into each expert’s prototype reservoir with sampling probability $w_k(\vz)$, 
thereby completing the initial population of the prototype memories without relying on the routing mechanism, 
and preliminarily constructing a stable domain representation space.
\textbf{2) Main Training.} 
After the warm-up stage, all expert parameters of current MoE are unfrozen and the routing mechanism is activated. 
At each training step, the routing signal $s_k(\mathbf{z})$ selects the Top-$k$ experts with the highest scores to participate in the forward computation, 
while the current domain embedding $\mathbf{z}$ is written into the prototype memory of the highest-scoring expert. 
The entire stage is optimized end-to-end under the standard cross-entropy loss:
\begin{equation}
    \mathcal{L}_{\mathrm{CE}}=-\sum_{l=1}^{L}\log P_{\theta}\left(y_l \mid \mathbf{H}_0, y_{<l}\right),
\end{equation}
where $L$ denotes the length of the target answer sequence.

\section{Experiments}

\subsection{Experimental Setup}
\textbf{Datasets.} 
We evaluate the proposed method on 10 representative Video Question Answering (VidQA) datasets, including MSVD-QA \cite{xu2017video}, MSRVTT-QA \cite{xu2017video}, TrafficQA \cite{xu2021sutd}, PororoQA \cite{kim2017deepstory}, SVQA \cite{song2018explore}, three subsets from FunQA \cite{xie2024funqa} (MagicQA, HumorQA, and CreativeQA), and two subsets from the Perception Test benchmark \cite{patraucean2023perception} (PhysicsQA and AbstractionQA). These datasets together form a multi-dimensional evaluation benchmark. The details of the datasets are provided in Appendix.

\noindent \textbf{Evaluation Metrics.} 
We train the model sequentially on the continuously arriving datasets. After completing training on the final dataset, we evaluate the model on the test sets of all previously seen datasets. During evaluation, we assume that the dataset identity of each test sample is unknown. To comprehensively assess the model performance, we primarily adopt the average accuracy $A_t$ to measure its overall performance, and introduce backward forgetting $BWF_t$ to quantify the degree of forgetting on previously learned knowledge. This enables a deeper 
analysis of the trade-off between mitigating catastrophic forgetting and maintaining stability. 
Formally,
\begin{equation}
A_t = \frac{1}{t} \sum_{i=1}^{t} R_{t,i}, \hspace{5pt} BWF_t = \frac{1}{t-1} \sum_{i=1}^{t-1} (R_{i,i} - R_{t,i})
\end{equation}
where $R_{i,j}$ denotes the test accuracy on the $j$-th dataset after the model has been trained on the $i$-th dataset.

\begin{figure}[t]
    \centering
    \includegraphics[width=0.99\linewidth]{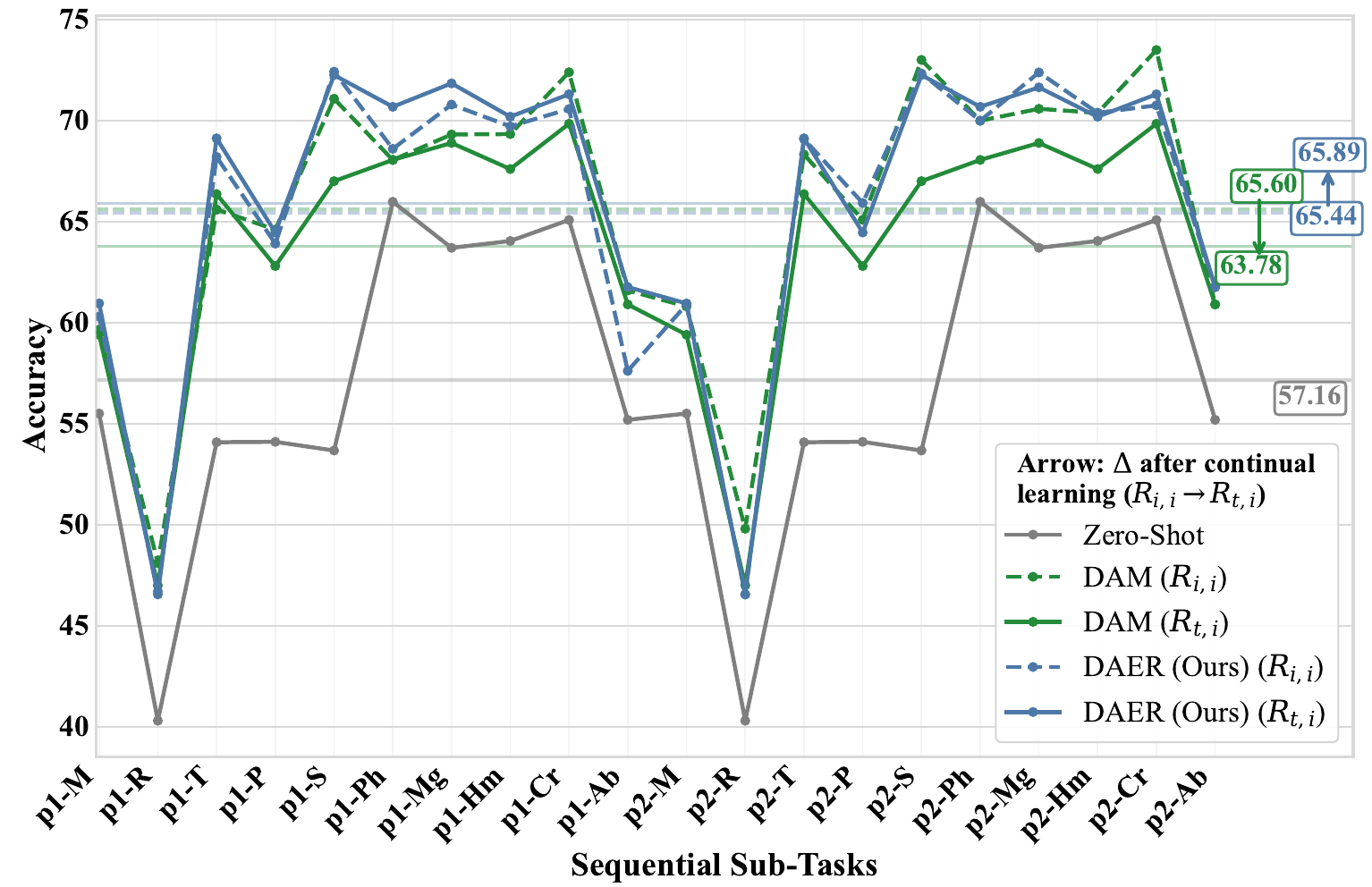}
    \caption{\textbf{Results under the 20-task continual learning setting.}
Each original domain is further divided into two sequential sub-tasks, yielding a 20-task stream.
}\label{fig:20tasks}
\end{figure}

\begin{table*}[t]
\centering
\caption{Performance comparison of different methods on ten tasks based on VideoLLaMA3.}
\label{tab:videollama3_method_comparison}
\begin{tabular}{c|cccccccccc|cc}
\toprule
Method & MSVD & MSRVTT & Traffic & Pororo & SVQA & Physics & Magic & Humor & Creative & Abstraction & Avg. & \lat{BWF$\downarrow$} \\
\midrule
Zero-Shot & 58.25 & 42.55 & 52.36 & 59.39 & 61.67 & 63.62 & 64.44 & 65.48 & 66.82 & 47.06 & 58.16 & -- \\
DAM & \textbf{63.80} & \textbf{50.70} & 58.80 & 64.05 & \textbf{73.67} & 62.93 & 65.93 & 65.38 & 68.74 & 45.67 & 61.97 & \lat{1.93} \\
\rowcolor{lightpurple} DAER (Ours) & 63.75 & 47.20 & \textbf{67.76} & \textbf{67.32} & 73.33 & \textbf{65.98} & \textbf{68.25} & \textbf{67.69} & \textbf{68.83} & \textbf{47.75} & \textbf{63.79} & \lat{\textbf{0.02}} \\
\bottomrule
\end{tabular}
\label{videollama}
\end{table*}

\begin{figure*}[t]
    \centering
    \includegraphics[width=0.99\linewidth]{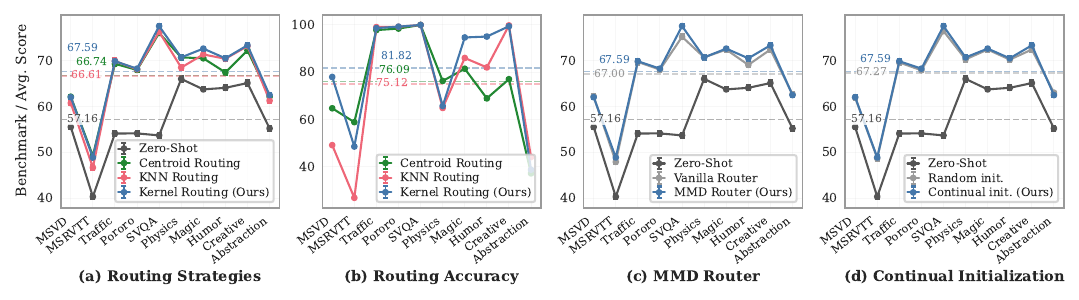}
    \caption{Ablation study on four important components.}
    \label{fig:ablation}
\end{figure*}

\begin{figure*}[t]
    \centering
    \includegraphics[width=0.99\linewidth]{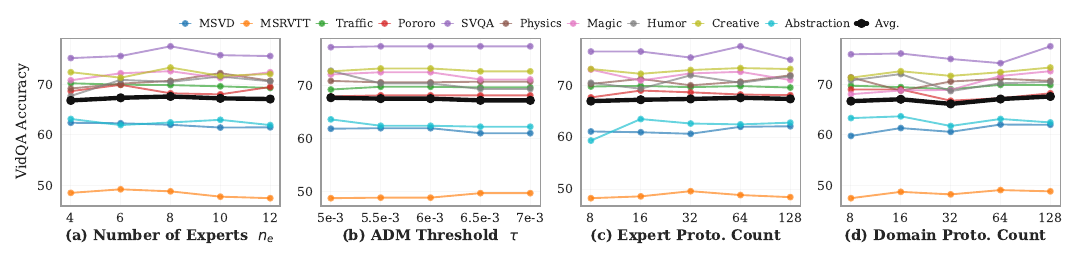}
    \caption{Sensitivity study on four critical hyper-parameters.}
    \label{fig:hyper}
\end{figure*}

\subsection{Main Results}
\noindent \textbf{Comparison with State-of-the-Art Methods.} \hspace{3pt}
Table~\ref{tab:main_results} reports the main results under the continual learning setting on InternVideo2.5-8B, where the model is trained over ten domains sequentially. Overall, DAER achieves the best average accuracy of $67.59\%$, outperforming the strongest prior continual baseline, DAM \cite{cheng2024dam}, by $3.21$ points ($67.59$ vs.\ $64.38$), while also attaining a negative backward forgetting score of $-0.14$, in contrast to the positive forgetting observed in DAM ($1.81$), MoE-Adapters \cite{yu2024boosting} ($3.43$), and LwF \cite{li2017learning} ($4.24$). Compared with other parameter-efficient adaptation strategies, DAER consistently performs better than prompt-based methods such as L2P \cite{wang2022learning} ($60.59\%$), CODA-Prompt \cite{smith2023coda} ($61.06\%$), and S-Prompts \cite{wang2022s} ($63.14\%$), as well as adapter-based methods including MoE-Adapters ($61.57\%$) and DAM ($64.38\%$), indicating that shared prompting or shared adapter spaces are insufficient to cope with the severe domain shifts in continual VidQA. A closer look at the per-domain results further highlights the advantage of DAER. The gains are especially pronounced on domains with stronger reasoning or stylistic shifts, such as SVQA, Creative, and Abstraction, where DAER improves over DAM by $6.25$, $3.29$, and $5.54$ points, respectively. 
Moreover, although the frozen Video-MLLM already provides a reasonable zero-shot average accuracy of $57.16\%$, continual adaptation with DAER improves this result by more than $10$ points, demonstrating that DAER not only adapts effectively to newly arriving domains but also preserves previously learned knowledge.

\vspace{3pt}
\noindent \textbf{Generalizing to 20 Tasks.}\hspace{5pt}
To further evaluate the robustness of the proposed method under a more challenging continual learning scenario, we consider a 20-task setting in which each original domain is split into two sequential sub-tasks.
As shown in Fig.~\ref{fig:20tasks}, DAER consistently achieves the strongest overall performance, obtaining an average accuracy of $65.89\%$, which outperforms DAM by $2.11$ points and the zero-shot baseline by $8.73$ points. At the same time, DAER exhibits substantially better stability, achieving a slightly negative BWF of $-0.47$, whereas DAM still suffers from noticeable forgetting with a BWF of $1.91$, indicating that previously learned knowledge is better preserved throughout the longer task sequence. Moreover, the task-wise curves show that DAER maintains strong performance not only on later tasks but also on earlier ones, yielding more consistent accuracy and avoiding the larger performance drops observed in DAM after multiple sequential updates. 

\vspace{3pt}
\noindent \textbf{Generalization to Other Video-MLLMs.} \hspace{5pt}
Table~\ref{tab:videollama3_method_comparison} evaluates the proposed method on VideoLLaMA3 to verify its generalization beyond InternVideo2.5. DAER again achieves the best overall performance, improving the average accuracy from $61.97\%$ with DAM to $63.79\%$, while reducing backward forgetting from $1.93$ to $0.02$. Compared with the zero-shot baseline, DAER also yields a substantial gain of more than $5$ points in average accuracy ($63.79\%$ vs.\ $58.16\%$). Although DAM is slightly better on a few domains such as MSRVTT and SVQA, DAER performs strongly on most others.

\vspace{-8pt}
\subsection{Model Analysis}
\noindent \textbf{Ablation Studies.} \hspace{3pt}
We conduct ablation studies to evaluate the main components of DAER. As shown in Fig.~\ref{fig:ablation}(a), the proposed kernel routing achieves the best overall performance, reaching an average score of $67.59$, which is higher than centroid routing ($66.74$) and KNN routing ($66.61$), and substantially better than the zero-shot baseline ($57.16$).
Fig.~\ref{fig:ablation}(b) further shows that kernel routing also attains the highest routing accuracy across most tasks, confirming that the gains of DAER are closely tied to improved expert assignment quality. In Fig.~\ref{fig:ablation}(c), replacing a vanilla router with the proposed MMD router improves the average score from $67.00$ to $67.59$, demonstrating that MMD offers a more effective distribution-aware similarity measure for expert selection under sequential domain shifts. Finally, Fig.~\ref{fig:ablation}(d) shows that continual initialization yields a higher average score than random initialization ($67.59$ vs.\ $67.27$) and produces more stable performance over the task stream, suggesting that inheriting parameters from the previously learned expert pool provides a better starting point for adapting to new domains while preserving task-level parameter isolation. 

\begin{table}[t]
\centering
\caption{Efficiency comparison with different methods.}
\setlength{\tabcolsep}{1pt}
\resizebox{0.45\textwidth}{!}{%
\begin{tabular}{lcccc}
\toprule
Method & Train.Params.(M)$\downarrow$ & GFLOPs$\downarrow$ & Latency (s)$\downarrow$ & Avg. Acc. (\%)$\uparrow$ \\
\midrule
S-Prompts      & 3.28   & 27.69  & 0.87 & 63.14 \\
CODA-Prompt    & 26.28  & 28.21  & 0.87 & 61.06 \\
MoE-Adapters   & 91.77  & 36.62  & 1.22 & 61.57 \\
\rowcolor{gray}
DAM            & 302.30 & 568.05 & 1.46 & 64.38 \\
\rowcolor{lightpurple}
DAER (Ours)    & 139.54 & 41.63  & 1.33 & 67.59 \\
$\Delta$ & \textcolor{red}{-53.84\%} & \textcolor{red}{-92.67\%} & \textcolor{red}{-8.90\%} & \textcolor{red}{+3.21} \\
\bottomrule
\end{tabular}}
\label{tab:efficiency_comparison}
\end{table}

\begin{table}[t]
\centering
\caption{Performance under different domain orders. V: MSVD, R: MSRVTT, T: Traffic, P: Pororo, S: SVQA, H: Physics, M: Magic, U: Humor, C: Creative, A: Abstraction.}
\label{tab:domain_order}
\begin{tabular}{cccccccccc|cc}
\toprule
\multicolumn{10}{c|}{Domain Order} & Avg. & \lat{BWF$\downarrow$} \\
\cmidrule(lr){1-12}
V & R & T & P & S & H & M & U & C & A & 67.59 & \lat{-0.14} \\
T & P & V & C & H & R & U & S & A & M & 67.40 & \lat{-0.06} \\
H & U & V & A & T & M & R & P & C & S & 67.69 & \lat{-0.21} \\
\bottomrule
\end{tabular}
\end{table}

\begin{figure}[t]
    \centering
    \includegraphics[width=0.99\linewidth]{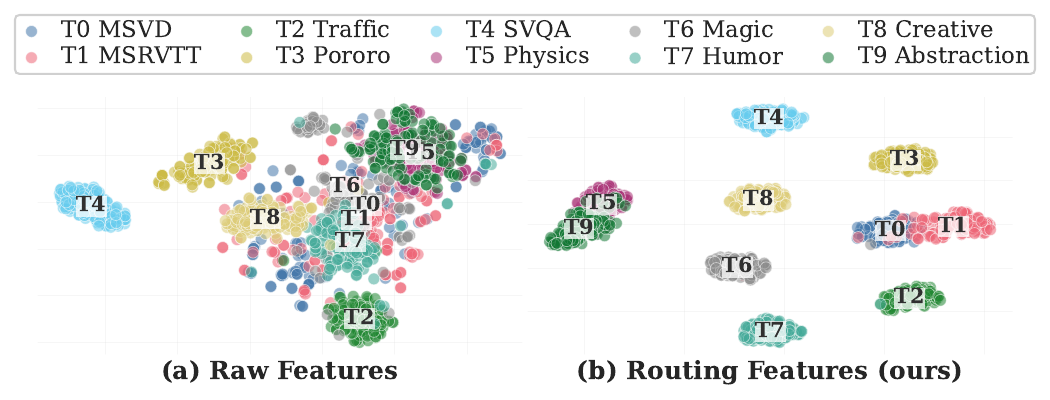}
    \caption{Visualization of domain routing features.}
    \label{fig:vis_routing}
\end{figure}

\vspace{3pt}
\noindent \textbf{Hyper-parameter Sensitivity.} \hspace{3pt}
We further study the sensitivity of DAER to several key hyper-parameters, including the number of experts $n_e$, the adaptive domain merging threshold $\tau$, the number of expert prototypes, and the number of domain prototypes, as shown in Fig.~\ref{fig:hyper}. Overall, DAER is fairly stable across a broad range of settings, indicating that the proposed framework is not overly sensitive to hyper-parameter choices. In particular, the average performance reaches its best level when using $n_e=8$ experts, $\tau=0.006$, and $64$ expert prototypes. The results also remain relatively consistent under different domain prototype counts, suggesting that the inter-domain routing mechanism is robust to moderate changes in prototype memory size. 

\vspace{3pt}
\noindent \textbf{Domain Order Robustness.} \hspace{3pt}
To evaluate robustness to domain arrival orders, Table~\ref{tab:domain_order} reports DAER performance under three randomly sampled domain arrival orders. DAER achieves stable results with an average accuracy of $67.56 \pm 0.15\%$, demonstrating that its effectiveness does not depend on a specific domain order.

\vspace{3pt}
\noindent \textbf{Efficiency Analysis.} \hspace{3pt}
Table~\ref{tab:efficiency_comparison} reports the efficiency and performance comparison of different methods. Although DAM is the strongest prior baseline in accuracy ($64.38\%$), it requires $302.30$M trainable parameters, $568.05$ GFLOPs, and $1.46$s latency. In contrast, DAER achieves a higher average accuracy of $67.59\%$ with only $139.54$M trainable parameters, $41.63$ GFLOPs, and $1.33$s latency, yielding improvements of $3.21$ percentage points in average accuracy while reducing trainable parameters, GFLOPs, and latency by $53.84\%$, $92.67\%$, and $8.90\%$, respectively. Compared with lighter prompt-based baselines, although DAER introduces more adaptation parameters than lightweight prompt-based methods, it achieves a better trade-off between efficiency and continual adaptation performance. Overall, these results demonstrate that DAER achieves both higher effectiveness and better efficiency than prior continual learning methods.

\begin{figure}[t]
    \centering
    \includegraphics[width=0.99\linewidth]{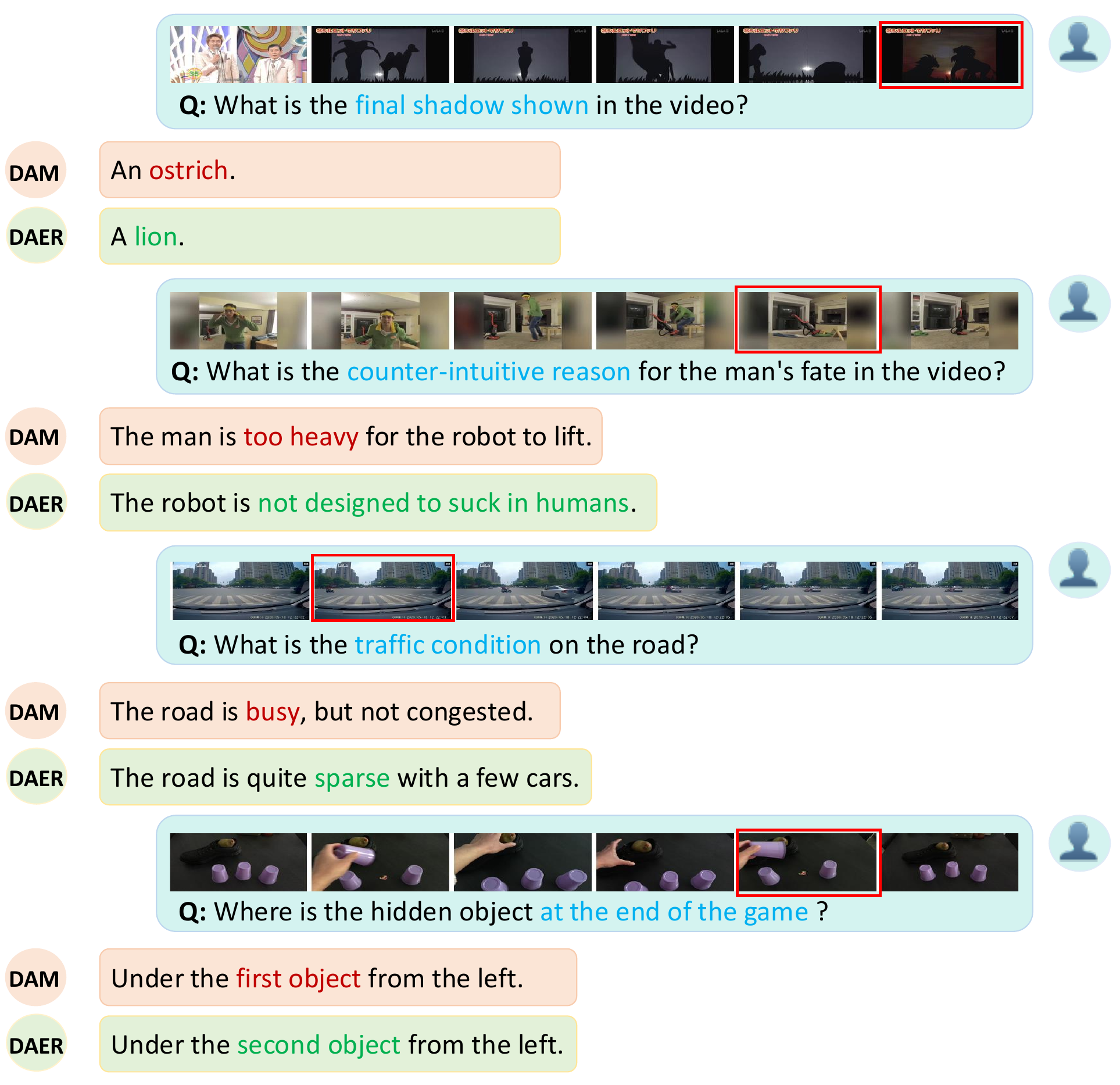}
    \caption{Qualitative comparison between DAM and the proposed DAER on representative VidQA examples. }
    \label{fig:vqa}
\end{figure}

\subsection{Qualitative Analysis}
To better understand the proposed routing mechanism, Fig.~\ref{fig:vis_routing} compares the raw multimodal features from the frozen backbone with the learned routing features from our domain router, where each point denotes a sample and colors indicate different domains. As shown in Fig.~\ref{fig:vis_routing}(a), the raw features exhibit substantial overlap across domains, suggesting that the frozen backbone alone does not provide sufficiently discriminative structure for reliable expert selection. In contrast, the learned routing features in Fig.~\ref{fig:vis_routing}(b) form much more compact and well-separated clusters, with tighter intra-domain grouping and larger inter-domain margins. 
Fig.~\ref{fig:vqa} provides qualitative comparisons between DAER and DAM on challenging VidQA examples. DAER achieves more accurate answers on cases requiring visual reasoning and temporal understanding, while DAM is more prone to misleading visual cues.

\section{Conclusion}
In this study, we investigate the problem of domain-incremental continual learning for video question answering and propose the DAER framework. The proposed method achieves parameter-level decoupling through domain-isolated lightweight expert modules, and combines MMD-based distribution-aware routing with a kernel-similarity-based task identification mechanism to enable adaptive recovery of task-specific states without requiring explicit task identities. Experimental results show that our method achieves state-of-the-art overall continual learning performance on the ten-domain benchmark.

\section*{Acknowledgments}
This work is supported by the National Natural Science Foundation of China under Grant No. 62502429, and the Zhejiang Key Laboratory Project (2024E10001).

\clearpage
\noindent\newpage

\bibliographystyle{ACM-Reference-Format}
\bibliography{main}

\clearpage

\appendix
\section*{Appendix}

\begin{table*}[t]
\centering
\small
\renewcommand{\arraystretch}{1.08}
\setlength{\tabcolsep}{4.2pt}
\caption{Comparison between oracle task-id inference and free-routing inference.}
\label{tab:taskid_analysis}
\resizebox{\textwidth}{!}{
\begin{tabular}{llccccccccccccc}
\toprule
\multirow{2}{*}{Method} 
& \multirow{2}{*}{Inference Setting} 
& \multicolumn{10}{c}{Downstream VidQA Accuracy (\%)} 
& \multirow{2}{*}{Avg.} 
& \multirow{2}{*}{\lat{BWF$\downarrow$}} 
& \multirow{2}{*}{Gap$\downarrow$} \\
\cmidrule(lr){3-12}
& & MSVD & MSRVTT & Traffic & Pororo & SVQA & Physics & Magic & Humor & Creative & Abstraction 
& & & \\
\midrule
\multirow{2}{*}{DAM}
& Oracle Task-ID 
& \textbf{61.35} & \textbf{51.85} & \textbf{69.12} & \textbf{66.41} & \textbf{72.00} & \textbf{70.26} & \textbf{71.01} & \textbf{73.85} & \textbf{72.03} & \textbf{62.28}

& \textbf{67.02} & -- & \multirow{2}{*}{2.64} \\
& Free Routing
& 59.95 & 48.05 & 67.84 & 64.95 & 71.25 & 67.36 & 68.89 & 68.56 & 70.02 & 56.92 
& 64.38 & \lat{1.81} & \\
\midrule
\multirow{2}{*}{DAER (Ours)}
& Oracle Task-ID
& \textbf{62.20} & \textbf{52.30} & 69.60 & \textbf{68.29} & \textbf{77.50} & \textbf{70.95} & 71.96 & \textbf{73.56} & 73.13 & 61.76 
& \textbf{68.13} & -- & \multirow{2}{*}{0.54} \\
& Free Routing 
& 62.00 & 48.85 & \textbf{69.84} & 68.22 & \textbf{77.50} & 70.68 & \textbf{72.59} & 70.48 & \textbf{73.31} & \textbf{62.46} 
& 67.59 & \textbf{\lat{-0.14}}& \\
\bottomrule
\end{tabular}
}
\end{table*}

\section{More Implementation Details} 

\noindent \textbf{Technical Details of DAER.}
We select InternVideo2.5 \cite{wang2025internvideo2} and VideoLLaMA3 \cite{zhang2025videollama} as two representative Video-MLLMs to demonstrate the generality and effectiveness of our method across different architectures. Specifically, InternVideo2.5 emphasizes the joint modeling of long-range video context and fine-grained visual information, enabling better capture of long-range temporal dependencies and improved understanding of complex dynamic content. In contrast, VideoLLaMA3 adopts a vision-centric training paradigm and architectural design, demonstrating strong overall performance on both image and video understanding tasks, with particular emphasis on high-quality visual representation learning. Experiments on these two representative backbone models further demonstrate that our method does not rely on a specific network architecture or a single video modeling mechanism, but can instead serve as a general continual adaptation framework that generalizes well across different Video-MLLM architectures.

Both InternVideo2.5 and VideoLLaMA3 follow a common architectural paradigm that couples a vision encoder with a large language model. Considering that existing parameter-efficient fine-tuning methods for vision-language models are typically applied mainly to the language model component, while the vision encoder is generally kept frozen, we further extend this setting to the Video-MLLM scenario. Specifically, we adapt only the large language model component by inserting the proposed lightweight adaptation module, DAER, in parallel to the feed-forward network of each Transformer block. For each task, we configure 8 experts, where each expert adopts a LoRA-like lightweight bottleneck structure with a unified rank of 32, and a Top-$k$ =  routing strategy is employed. The batch size is set to 12 for MSVD-QA \cite{xu2017video} and MSRVTT-QA \cite{xu2017video}, and to 8 for all the other tasks.

We adopt a two-stage training strategy. During the warm-up stage, we use only 30\% of the training data from the current task for routing initialization, with the learning rate set to 4e-4. During the main training stage, we use AdamW as the optimizer with a weight decay of 0.01, together with a learning rate schedule that combines linear warm-up and cosine annealing, where the base learning rate is set to 2e-4. Specifically, the learning rate is first linearly increased to its peak value during the first 3\% of the total training steps, and then gradually decayed to 0 following a cosine annealing schedule over the remaining training steps.

\noindent \textbf{Continual Learning Baselines.}
We implement a range of classical continual learning baselines and compare them with the proposed method. In general, these baselines can be grouped into three categories: regularization-based methods, including LwF \cite{li2017learning} and EWC \cite{kirkpatrick2017overcoming}; prompt-based methods, including L2P \cite{wang2022learning}, CODA-Prompt \cite{smith2023coda}, and S-Prompts \cite{wang2022s}; and adapter-based methods, including DAM \cite{cheng2024dam} and MoE-Adapters \cite{yu2024boosting}. To ensure a fair comparison across different methods, all approaches are implemented and evaluated on the same pretrained backbone. Considering that existing parameter-efficient fine-tuning methods for VLMs are typically applied to the language model component, while the vision encoder is generally kept frozen, we follow the same setting in this work.

\begin{itemize}

    \item \textbf{S-Prompts} \cite{wang2022s}. This method is specifically designed for domain-incremental continual learning. By assigning independent prompt parameters to different domains, it achieves parameter isolation across domains, thereby effectively alleviating catastrophic forgetting caused by distribution shifts during continual adaptation. In our implementation, the prompt length is set to 50. We apply K-Means clustering to the feature representations of each task and retain 3 cluster centers for each task. During inference, we adopt a nearest-neighbor matching strategy to retrieve the cluster center that is closest to the current sample from all task centers, determine its corresponding task, and then select the associated prompt accordingly. The learning rate is set to 1e-2.
    
    \item \textbf{CODA-Prompt} \cite{smith2023coda}. To ensure a fair comparison, the prompt length of CODA-Prompt is set to be consistent with that of S-Prompts. For each new task, we adopt an orthogonal initialization strategy to initialize its prompt components, key vectors, and attention vectors. In addition, orthogonality constraints are introduced during training to promote the decoupling of prompt components across different tasks, thereby mitigating inter-task interference.
    
    \item \textbf{L2P} \cite{wang2022learning}. We adopt the same prompt setting as S-Prompts. During training, the representations in the prompt pool are continuously optimized for different tasks. During inference, the most relevant prompts are selected for forward computation based on the similarity between the current sample and the prompt keys.
    
    \item \textbf{MoE-Adapters} \cite{yu2024boosting}. We adopt a configuration with 8 experts for each MoE-Adapters layer and employ a Top-2 sparse routing strategy, such that only the two highest-scoring experts are activated during each forward pass. Each expert is implemented as a LoRA-like lightweight bottleneck with a rank of 32. The main learning rate is set to 2e-4, while the learning rate for DDAS is set to 2e-5. Following the implementation protocol of the original method, we dynamically freeze frequently used experts according to their usage frequency. Specifically, after training on each task, we freeze the expert that is selected most frequently for that task.
    
    \item \textbf{DAM} \cite{cheng2024dam}. Similar to S-Prompts, this method is also designed for the domain-incremental continual learning setting. We follow its original architecture by inserting adapters after both the self-attention module and the FFN module in each Transformer block. Each adapter consists of a down-projection linear layer and an up-projection linear layer, where the former maps the features to an intermediate hidden dimension with an 8$\times$ reduction ratio, and the latter projects them back to the original feature dimension. During training, a separate set of adapter parameters is learned for each task. During inference, the adapter parameters corresponding to the Top-$k$ most similar tasks are adaptively fused according to the similarity between the current sample feature and the feature centroid of each previously learned task, where $k = 2$ and the fusion temperature is set to 0.01. The learning rate is set to 1e-4.

    \item \textbf{LwF} \cite{li2017learning} and \textbf{EWC} \cite{kirkpatrick2017overcoming}. These two regularization-based continual learning methods were originally designed primarily for the full fine-tuning setting. Considering that our work adopts a parameter-efficient fine-tuning paradigm, we adapt these two methods to our parameter-efficient framework while preserving their original design principles as much as possible to ensure a fair comparison. Specifically, we apply the corresponding regularization constraints to the adapter parameters and update only the adapters, while keeping all other model parameters frozen. Meanwhile, all domains share a single set of adapter parameters.
    
\end{itemize}

\section{Dataset Descriptions}

We conduct continual learning experiments on ten VideoQA datasets, including MSVD-QA \cite{xu2017video}, MSRVTT-QA \cite{xu2017video}, TrafficQA \cite{xu2021sutd}, PororoQA \cite{kim2017deepstory}, SVQA \cite{song2018explore}, PhysicsQA \cite{patraucean2023perception}, MagicQA \cite{xie2024funqa}, HumorQA \cite{xie2024funqa}, CreativeQA \cite{xie2024funqa}, and AbstractionQA \cite{patraucean2023perception}. Below we provide a detailed description of the selected datasets:

\begin{itemize}
    \item \textbf{MSVD-QA} \cite{xu2017video} is an open-domain video question answering dataset whose videos are mainly drawn from daily-life scenarios. It covers information such as human actions, object attributes, and event processes, making it suitable for evaluating a model's ability to understand general video content.
    
    \item \textbf{MSRVTT-QA} \cite{xu2017video} contains more diverse video topics than MSVD-QA, covering a wide range of scenarios such as entertainment, sports, and daily life, thereby imposing higher demands on the model's video understanding and generalization abilities.
    
    \item \textbf{TrafficQA} \cite{xu2021sutd} consists of videos collected from real traffic accident scenes. The videos typically involve road environments, vehicle behaviors, interactions among traffic participants, and potential risk events, and are mainly used to evaluate the model's understanding and reasoning abilities in complex dynamic traffic scenarios.
    
    \item \textbf{PororoQA} \cite{kim2017deepstory} is derived from animated videos, and its questions mainly focus on character behaviors, plot development, and event relationships. Due to its strong narrative nature, this dataset is well suited for evaluating a model's ability to understand continuous storylines and character interactions.
    
    \item \textbf{SVQA} \cite{song2018explore} is a video question answering dataset constructed from synthetic video scenes. It typically focuses on object attributes, motion states, and numerical relationships, and can effectively assess a model's basic visual reasoning ability.
    
    \item \textbf{PhysicsQA} and \textbf{AbstractionQA} are both derived from Perception Test \cite{patraucean2023perception}. Perception Test is a diagnostic video benchmark for evaluating the perception and reasoning abilities of pretrained multimodal models. Through purposefully designed and filmed real-world videos, it systematically examines model performance across multiple capability dimensions, including memory, abstraction, physics, and semantics. Among them, PhysicsQA mainly focuses on physical phenomena and dynamic causal relationships, whereas AbstractionQA places greater emphasis on understanding high-level semantics and abstract concepts.
    
    \item \textbf{MagicQA}, \textbf{HumorQA}, and \textbf{CreativeQA} are all derived from FunQA \cite{xie2024funqa}. FunQA is a video question answering benchmark centered on surprising videos, aiming to evaluate models' ability to understand and reason about counter-intuitive video content. Among them, HumorQA mainly focuses on contrasts and unexpected events in humorous videos, emphasizing the understanding of humorous context and counter-intuitive events; CreativeQA focuses on videos with creative expressions or cleverly disguised content, placing more emphasis on understanding novel visual semantics and unconventional expressions; MagicQA centers on magic or visual illusion videos, and is used to evaluate the model's perception and reasoning abilities regarding seemingly impossible events and their counter-intuitive elements.
\end{itemize}

\begin{table*}[t]
\centering
\small
\renewcommand{\arraystretch}{1.08}
\setlength{\tabcolsep}{4.2pt}
\caption{Detailed Component Analysis of DAER.}
\label{tab:detailed_ablation}
\begin{tabular}{llcccccccccccc}
\toprule
\multirow{2}{*}{Category} & \multirow{2}{*}{Method}
& \multicolumn{10}{c}{Downstream VidQA Accuracy (\%)}
& \multirow{2}{*}{Avg.}
& \multirow{2}{*}{\lat{BWF$\downarrow$}} \\
\cmidrule(lr){3-12}
& & MSVD & MSRVTT & Traffic & Pororo & SVQA & Physics & Magic & Humor & Creative & Abstraction
& & \\
\midrule
\multirow{3}{*}{Merging}
& Random Merging
& 61.90 & 48.85 & 68.76 & 67.87 & 76.25 & 70.95 & 71.85 & 68.75 & 72.94 & 62.46
& 67.06 & \lat{0.10} \\
& ADM (Ours)
& \textbf{62.00} & 48.85 & \textbf{69.84} & \textbf{68.22} & \textbf{77.50} & \textbf{70.68} & \textbf{72.59} & 70.48 & \textbf{73.31} & \textbf{62.46}
& 67.59 & \textbf{\lat{-0.14}} \\
& No Merging
& 61.85 & \textbf{48.95} & 69.76 & \textbf{68.22} & \textbf{77.50} & 70.54 & 72.17 & \textbf{73.27} & 73.13 & 61.42
& \textbf{67.68} & \lat{0.06} \\
\midrule
\multirow{3}{*}{Kernel}
& Single-Gaussian
& 61.55 & 48.55 & 69.76 & 67.52 & 76.25 & 70.82 & 72.38 & 70.38 & 72.76 & 61.59
& 67.16 & \lat{0.22} \\
& Multi-Gaussian
& 62.00 & 48.55 & \textbf{69.84} & \textbf{68.29} & \textbf{77.58} & \textbf{70.68} & 72.49 & 70.29 & 72.94 & 62.11
& 67.48 & \lat{-0.09} \\
& Cosine (Ours)
& \textbf{62.00} & \textbf{48.85} & \textbf{69.84} & 68.22 & 77.50 & \textbf{70.68} & \textbf{72.59} & \textbf{70.48} & \textbf{73.31} & \textbf{62.46}
& \textbf{67.59} & \textbf{\lat{-0.14}} \\
\midrule
\multirow{2}{*}{Warm-up}
& w/o Warm-up
& 61.80 & \textbf{50.85} & 69.40 & 67.80 & 74.67 & 69.57 & 71.74 & \textbf{71.35} & 72.76 & 61.42
& 67.14 & \textbf{\lat{-0.23}} \\
& w/ Warm-up (Ours)
& \textbf{62.00} & 48.85 & \textbf{69.84} & \textbf{68.22} & \textbf{77.50} & \textbf{70.68} & \textbf{72.59} & 70.48 & \textbf{73.31} & \textbf{62.46}
& \textbf{67.59} & \lat{-0.14} \\
\bottomrule
\end{tabular}
\end{table*}

\section{Task-Free Inference Analysis}

To further supplement the analysis in the main text regarding inference without explicit task identity, we compare the model performance under two settings: oracle task-id inference and free-routing inference. Specifically, oracle task-id inference refers to the setting where the ground-truth task-id of each test sample is provided during inference. For DAM, this corresponds to directly using the adapter parameters trained for the corresponding task, without performing parameter merging; for our proposed DAER, it corresponds to directly invoking the expert parameters associated with the corresponding task under the condition that the true task-id is known. In contrast, free-routing inference refers to the setting where no ground-truth task-id is provided, and each method performs inference according to its original routing mechanism. 

It is worth noting that, as discussed in the main text, a key practical challenge in domain-incremental continual learning lies in the fact that explicit task identity is usually unavailable at test time. To address this issue, we introduce an inter-domain distribution-aware routing mechanism for task prediction; meanwhile, in our experimental setup, we consistently evaluate all methods under the setting where the task identity of each test sample is unknown. Based on this setting, Table~\ref{tab:taskid_analysis} further examines the performance degradation of the two methods when inference shifts from oracle task-id inference to free-routing inference. As shown in the results, both methods suffer some performance drop after removing the ground-truth task identity, but the extent of degradation differs substantially. Specifically, the average accuracy of DAM drops from 67.02 to 64.38, yielding a Gap of 2.64. In comparison, the average accuracy of DAER only decreases from 68.13 to 67.59, with a much smaller Gap of 0.54. Moreover, DAER not only achieves better average performance under oracle task-id inference, but also maintains a more stable BWF under free-routing inference. These results indicate that DAER is less dependent on explicit task identity, and that its distribution-aware routing mechanism enables more robust expert selection in the absence of task priors, thereby effectively mitigating the performance degradation caused by the absence of task identity.

\section{Merge Strategy Analysis}

As a further analysis, we compare different merging strategies to evaluate whether the proposed Adaptive Domain Merging (ADM) mechanism can reduce parameter redundancy while preserving continual learning performance. As shown in the Merging part of Table~\ref{tab:detailed_ablation}, ADM achieves the best overall trade-off among the three strategies. Specifically, compared with random merging, ADM improves the average accuracy from 67.06\% to 67.59\%, while reducing backward forgetting from 0.10 to -0.14. This indicates that indiscriminately merging task-specific parameters in a random manner can easily introduce additional cross-domain interference, making it difficult to effectively preserve previously acquired knowledge.

In contrast, the no-merging strategy preserves all task-specific parameters and therefore provides a strong parameter-unconstrained reference, but its parameter size grows linearly with the number of tasks. Compared with No Merging, ADM yields a slightly lower average accuracy (67.59\% vs.\ 67.68\%), but outperforms the no-merging strategy on several tasks, such as MSVD, Magic, and Creative. Moreover, ADM achieves a better BWF (-0.14 vs.\ 0.06), indicating that the proposed similarity-aware adaptive merging mechanism can better alleviate forgetting while maintaining competitive overall performance.

Overall, these results verify that the proposed method is able to exploit inter-task similarity in a more principled manner, thereby maintaining strong continual adaptation capability while controlling the growth of parameter size.

\section{Kernel Function Analysis for Task Routing}
Similarly, since we introduce kernel functions in Inter-Domain Distribution-Aware Routing to measure the similarity between the input sample and task prototypes, we further compare different kernel functions for task routing to analyze how the form of similarity measure affects routing quality and continual adaptation performance. Specifically, the Single-Gaussian Kernel adopts a single bandwidth parameter with $\sigma = 1$, while the Multi-Gaussian Kernel employs a multi-kernel combination in which the Gaussian bandwidths are set to $\sigma \in \{0.5, 1, 2, 4\}$, so as to better capture distribution discrepancies at different scales. As shown in the Kernel part of Table~\ref{tab:detailed_ablation}, the Cosine Kernel achieves the best overall performance among all compared methods, reaching an average accuracy of 67.59\% and the lowest backward forgetting of -0.14. In contrast, the Single-Gaussian Kernel only achieves an average accuracy of 67.16\% with a BWF of 0.22, indicating that a single Gaussian kernel has relatively limited distribution discrimination ability and is less capable of supporting more stable and accurate routing decisions. The Multi-Gaussian Kernel improves the average accuracy to 67.48\% and reduces the BWF to -0.09, suggesting that multi-kernel Gaussian similarity already provides stronger distribution characterization than the single-kernel design. Nevertheless, from the overall results, the Cosine Kernel still performs better on most tasks and achieves the best balance between accuracy and forgetting.

\section{Effectiveness Analysis of the Warm-up Stage}

The Warm-up part of Table~\ref{tab:detailed_ablation} validates the effectiveness of the warm-up stage in the proposed two-stage training strategy. According to our design, the warm-up stage is mainly introduced to initialize the prototype reservoirs and establish a more stable routing space before the full DAER routing mechanism is activated, thereby alleviating the cold-start instability of expert prototypes during the early stage of training. Without the warm-up strategy, we observe a clear imbalance in expert utilization during training, where the MoE gradually collapses into a degenerate state in which only a few experts remain consistently active. This phenomenon suggests that, in the absence of stable initialization, the routing mechanism struggles to form a reasonable division of labor among experts. In contrast, after introducing the warm-up stage, the average accuracy improves from 67.14\% to 67.59\%, indicating that a more stable early-stage initialization can effectively enhance the overall adaptation performance on continual VidQA tasks. More specifically, warm-up brings performance gains on the vast majority of datasets, with particularly notable improvement on SVQA. This suggests that warm-up helps establish a more reliable correspondence between experts and underlying data distributions at the beginning of training, thereby improving subsequent routing quality and expert specialization. Although performance drops are observed on MSRVTT and Humor, the overall trend still demonstrates that the warm-up stage provides a more robust initialization process for DAER, thereby enhancing its adaptability and robustness in evolving domains. This observation is also consistent with the design motivation of the proposed two-stage training strategy, which is intended to stabilize expert specialization during continual adaptation.

\begin{table}[t]
\centering
\small
\setlength{\tabcolsep}{5.5pt}
\renewcommand{\arraystretch}{1.08}
\caption{Generalization of DAER to language-guided object tracking and subsequent VidQA adaptation.}
\label{tab:lasot_rebuttal}
\begin{tabular}{@{}lcccc@{}}
\toprule
Setting & IoU & AUC &
VidQA Avg. & \lat{BWF $\downarrow$} \\
\midrule
Zero-Shot     & 3.76  & 4.53  & 57.16 & --    \\
After LaSOT   & 26.33 & 26.72 & --    & --    \\
After 3 tasks & 26.20 & 26.58 & 67.28 & \lat{0.41}  \\
w/o LaSOT     & --    & --    & 67.59 & \lat{-0.14} \\
\bottomrule
\end{tabular}
\end{table}

\section{Generalization to Other Video-Language Tasks} 
We further incorporate a subset of LaSOT~\cite{fan2019lasot}, a language-guided video object tracking benchmark, into the continual learning sequence to evaluate whether DAER can generalize beyond video question answering to other video-language tasks. Specifically, we insert LaSOT before the final three tasks in the original VidQA continual learning sequence and subsequently continue sequential training on these three VidQA tasks. The results are reported in Table~\ref{tab:lasot_rebuttal}. After adapting to LaSOT, DAER substantially improves its tracking performance over the zero-shot model, with the IoU and AUC scores increasing from 3.76 and 4.53 to 26.33 and 26.72, respectively. After subsequently learning three VidQA tasks, the tracking performance decreases only marginally to 26.20 IoU and 26.58 AUC, indicating that the acquired tracking capability is effectively retained despite continued adaptation to heterogeneous downstream tasks. Meanwhile, DAER achieves an average accuracy of 67.28\% and a BWF of only 0.41 across all VidQA domains, which is comparable to the setting without LaSOT. These results demonstrate that DAER can accommodate a video-language task with a substantially different output format while effectively preserving previously acquired knowledge and maintaining stable performance on subsequent VidQA domains. This further verifies the cross-task generalization, knowledge isolation, and resistance to catastrophic forgetting of DAER in more complex and heterogeneous continual learning scenarios.

\section{Efficiency and Scalability Analysis}
DAER is designed to maintain favorable scalability as the number of domains increases during continual learning. First, each domain-specific expert is implemented as a lightweight LoRA module, eliminating the need to maintain a separate copy of the full backbone model for each domain. In addition, the memory reservoir of each domain stores only 128 prototypes, resulting in limited additional parameter and storage overhead. Although the computational cost of inter-domain routing increases approximately linearly with the number of domains, our experiments show that prototype matching against a single candidate domain requires only about 0.538\,ms in the 10-domain continual learning sequence. Therefore, even when the number of domains is extended to 100, the total routing latency is estimated to be only about 53.8\,ms, corresponding to approximately 4.0\% of the overall inference latency of 1330\,ms under the 10-domain setting. This indicates that the prototype-based inter-domain routing mechanism introduces only minor additional computational overhead, even for substantially longer domain sequences. Overall, DAER exhibits favorable scalability in terms of parameter growth, prototype storage, and inference latency, making it suitable for continual Video-MLLM adaptation over long and evolving domain sequences.

\end{document}